\documentclass[11pt,a4paper]{article}
\usepackage[hyperref]{emnlp2020}
\usepackage{times}
\usepackage{latexsym}

\usepackage{microtype}

\aclfinalcopy % Uncomment this line for the final submission
\def\aclpaperid{2764} %  Enter the acl Paper ID here

\usepackage{amsmath, amsfonts, amssymb}
\usepackage{bm}
\usepackage{graphicx}
\usepackage{booktabs}
\usepackage{multirow}

\usepackage{array}
\newcolumntype{L}{>{$}l<{$}}
\newcolumntype{C}{>{$}c<{$}}
\newcolumntype{R}{>{$}r<{$}}

\title{Neural Topic Modeling by Incorporating Document Relationship Graph}

\author{Deyu Zhou\thanks{\; Corresponding author.} \quad Xuemeng Hu \quad Rui Wang \\
  School of Computer Science and Engineering, Key Laboratory of Computer Network \\
  and Information Integration, Ministry of Education, Southeast University, China \\
  {\tt \{d.zhou,xuemenghu,rui\_wang\}@seu.edu.cn} \\}

\date{}

\begin{document}
\maketitle
\begin{abstract}
  Graph Neural Networks (GNNs)
  that capture the relationships between graph nodes via message passing
  have been a hot research direction
  in the natural language processing community.
  In this paper, we propose Graph Topic Model (GTM), a GNN based neural topic model
  that represents a corpus as a document relationship graph.
  Documents and words in the corpus become nodes in the graph and
  are connected based on document-word co-occurrences.
  By introducing the graph structure,
  the relationships between documents are established through their shared words
  and thus the topical representation of a document is enriched by
  aggregating information from its neighboring nodes using graph convolution.
  Extensive experiments on three datasets were conducted
  and the results demonstrate the effectiveness of the proposed approach.
\end{abstract}

\section{Introduction}

Probabilistic topic models \citep{Blei2012PTM} are tools for discovering main themes from large corpora.
The popular Latent Dirichlet Allocation (LDA) \citep{Blei2003LDA}
and its variants \citep{lin2009jst, zhao-etal-2010-jointly, zhou-etal-2014-simple}
are effective in extracting coherent topics in an interpretable manner,
but usually at the cost of designing sophisticated and model-specific learning algorithm.
Recently, neural topic modeling that utilizes neural-network-based black-box inference
has been the main research direction in this field.
Notably,
NVDM
\citep{miao2016nvdm} employs
variational autoencoder (VAE) \citep{kingma2013vae} to model topic inference and document generation.
Specifically, NVDM consists of an encoder inferring topics from documents
and a decoder generating documents from topics,
where the latent topics are constrained by a Gaussian prior.
\citet{srivastava2017prodlda} argued that Dirichlet distribution is a more appropriate prior for topic modeling
than Gaussian in NVDM and proposed ProdLDA that approximates the Dirichlet prior with logistic normal.
There are also attempts that directly enforced a Dirichlet prior on the document topics.
W-LDA \citep{nan-etal-2019-topic} models topics
in the Wasserstein autoencoders \citep{tolstikhin2017wae} framework and
achieves distribution matching by minimizing their Maximum Mean Discrepancy (MMD) \citep{gretton2012mmd},
while adversarial topic model \citep{wang2019atm, wang-etal-2019-open, wang-etal-2020-neural}
directly generates documents from the Dirichlet prior
and such a process is adversarially trained with a discriminator
under the framework of Generative Adversarial Network (GAN) \citep{goodfellow2014gan}.

Recently, due to the effectiveness of Graph Neural Networks (GNNs)
\citep{li2015ggnn, kipf2016semi, zhou2018graph}
in embedding graph structures,
there is a surge of interests of applying GNN to natural language processing tasks
\citep{yasunaga-etal-2017-graph, song-etal-2018-graph, Yao2019textgcn}.
For example, GraphBTM \citep{zhu-etal-2018-graphbtm} is a neural topic model
that incorporates the graph representation of a document
to capture biterm co-occurrences in the document.
To construct the graph, a sliding window over the document is employed and all word pairs in the window are connected.

A limitation of GraphBTM is that
only word relationships are considered while ignoring document relationships.
Since a topic is possessed by a subset of documents in the corpus,
we believe that the topical neighborhood of a document, i.e., documents with similar topics,
would help determine the topics of a document.
To this end, we propose Graph Topic Model (GTM),
a neural topic model that a corpus is represented as a document relationship graph
where documents and words in the corpus are nodes
and they are connected based on document-word co-occurrences.
In GTM, the topical representation of a document node is
aggregated from its multi-hop neighborhood, including both document and word nodes,
using Graph Convolutional Network (GCN) \citep{kipf2016semi}.
As GCN is able to capture high-order neighborhood relationships,
GTM is essentially capable of modeling both word-word and doc-doc relationships.
In specific, the relationships between relevant documents are established by their shared words,
which is desirable for topic modeling as
documents belonging to one topic typically have similar word distributions.

The main contributions of the paper are:
\begin{itemize}
  \item We propose GTM, a novel topic model that incorporates document relationship graph to
        enrich document and word representations.
  \item We extensively experimented on three datasets and the results demonstrate the effectiveness of the proposed approach.
\end{itemize}

\section {Graph Topic Model}\label{methodology}

\subsection {Graph Representation of the Corpus}
We represent the whole corpus $\mathcal{D}$ with an undirected graph $\mathcal{G}=(\mathcal{N}, \mathcal{E})$,
where $\mathcal{N}$ and $\mathcal{E}$ are nodes and edges in the graph respectively.
To model both words and documents, each of them is represented as a node $n_i \in \mathcal{N}$,
which gives rise to $N=V+D$ nodes in total,
where $V$ is the size of vocabulary $\mathcal{V}$  and $D$ is the number of documents in corpus $\mathcal{D}$.
An edge $(n_i, n_j)$ indicates the relevance of node $n_i$ and $n_j$, whose weight is determined by
\begin{equation}\label{eq:connectivity}
  A_{i,j} =
  \begin{cases}
    \text{TF-IDF}_{ij}, & i \in \mathcal{D} \text{ and } j \in \mathcal{V} \\
    \text{TF-IDF}_{ji}, & i \in \mathcal{V} \text{ and } j \in \mathcal{D} \\
    1,                  & i=j                                              \\
    0,                  & \text{otherwise}
  \end{cases}
\end{equation}
where $\bm{A}$ is the adjacency matrix of $\mathcal{G}$ and
TF-IDF$_{ij}$ denotes the max-normalized TF-IDF (Term Frequency–Inverse Document Frequency) weight of word $j$ in document $i$.
Besides self-connections, we only apply positive weights to edges between documents and words,
while rely on the model to capture higher-order relationships, e.g. doc-doc and word-word relationships,
by applying graph convolutions on graph $\mathcal{G}$.

\subsection{Model Architecture}

\begin{figure}[t]
  \centering
  \resizebox{\columnwidth}{!}{
    \includegraphics[]{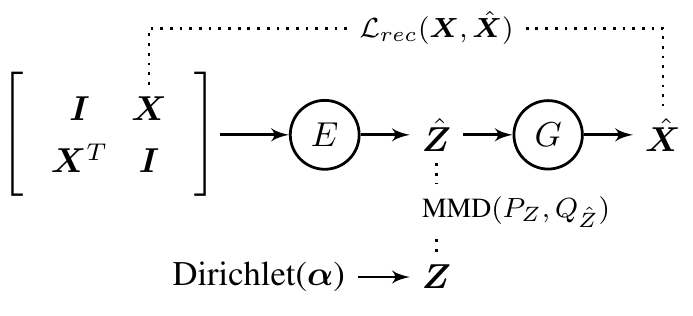}
  }
  \caption{The framework of GTM.
    Circles denote neural networks.
    $\bm{X}$, $\bm{I}$, $\hat{\bm{Z}}$, $\hat{\bm{X}}$, $\bm{Z}$ are
    the TF-IDF matrix of the corpus, an identity matrix, latent topics of all documents, reconstructed word weights and
    topic distributions drawn from the Dirichlet prior respectively.
    $\mathcal{L}_{rec}(\bm{X}, \hat{\bm{X}})$ and $\text{MMD}(P_Z, Q_Z)$ are training objectives.
  }
  \label{fig:framework}
\end{figure}

The proposed GTM consists of an encoder $E$ and a decoder $G$.
The framework is shown in Figure \ref{fig:framework}, and we detail the architecture in the following.

The encoder network $E$ maps nodes in $\mathcal{G}$ to their topic distributions
by iteratively applying graph convolution to the node features.
Following \citep{kipf2016semi}, the layer-wise propagation rule of the graph convolution
at layer $l+1 \in [1, L]$ is defined as
\begin{equation}\label{eq:gcn}
  \bm{H}^{(l+1)} = \sigma(\bm{D}^{-\frac{1}{2}} \bm{A} \bm{D}^{-\frac{1}{2}} \bm{H}^{{(l)}} \bm{W}^{(l)})
\end{equation}
where $\bm{A} \in \mathbb{R}^{N \times N}$ is the adjacency matrix of $\mathcal{G}$, $D_{ii}=\sum_j A_{ij}$,
$\bm{W}^{(l)} \in \mathbb{R}^{d^{(l)} \times d^{(l+1)}}$ is a layer-specific weight matrix
where $d^{(l)}$ is the output size of layer $l$,
and $\sigma$ denotes an activation function that is LeakyReLU \citep{maas2013leaky} in this paper.
$\bm{H}^{(l)} \in \mathbb{R}^{N \times d^{(l)}}$ is the activations of all nodes at layer $l$
and $\bm{H}^{(0)}_i$ is the embedding of node $i$.

At each encoder layer,
what the graph convolution does is aggregating node features from a node's first-order neighborhood,
which consequently enlarges the receptive field of the central node and
enables the information propagation between relevant nodes.
After successively applying $L$ graph convolution layers,
the encoding of a node essentially involves its $L^{\text{th}}$-order neighborhood.
With $L \geq 2$, doc-doc and word-word relationships are naturally captured in the topic inference process.

We also add a batch normalization \citep{Ioffe2015bn} after each graph convolution.
After the graph encoding,
a softmax is further applied to the node features of a document
to produce a multinomial topic distribution $\hat{\bm{z}} \in \mathbb{R}^K$,
where $K$ is the topic number.

Based on the inferred topic distribution $\hat{\bm{z}}$,
the decoder network $G$ tries to restore the original document representations.
To achieve this goal, we employ a 2-layer MLP with LeakyReLU activation and batch normalization in the first layer.
The output of the MLP decoder is then softmax-normalized
to generate a word distribution $\hat{\bm{x}} \in \mathbb{R}^V$.

The decoder is also used to interpret topics.
In this case, we feed to the decoder an identity matrix $\bm{I} \in \mathbb{R}^{K \times K}$,
and the decoder output $G(\bm{I})_i$ is the word distribution of the $i$-th topic.

\subsection{Training Objective}
Based on the Wasserstein Autoencoder \citep{tolstikhin2017wae} framework,
the training objective of GTM is to minimize the document reconstruction loss
when the latent topic space is constrained by a prior distribution.
The reconstruction loss is defined as
\begin{equation}
  \mathcal{L}_{rec}(\bm{X}, \hat{\bm{X}}) = -\mathbb{E} (\bm{x} \log{\hat{\bm{x}}}),
\end{equation}
where $\bm{x}$ denotes the TF-IDF of a document and
$\hat{\bm{x}}$ is the reconstructed word distribution corresponding to $\bm{x}$.
we use TF-IDF as the reconstruction target since TF-IDF basically preserves the relative importance of words
and reduces some background noise that may hurt topic modeling, e.g., stop words.

We impose a Dirichlet prior, the conjugate prior of the multinomial distribution,
to the latent topic distributions.
Following W-LDA \citep{nan-etal-2019-topic},
we achieve this goal by minimizing the Maximum Mean Discrepancy (MMD) \citep{gretton2012mmd}
between the distribution $Q_{\hat{Z}}$ of inferred topic distributions $\hat{\bm{z}}$
and the Dirichlet prior $P_Z$ from which we draw multinomial noises $\bm{z}$:
\begingroup\makeatletter\def\f@size{9}\check@mathfonts
\begin{align}
   & \text{MMD}(P_Z, Q_{\hat{Z}}) = \frac{1}{m(m-1)}\sum_{i \neq j} k(\bm{z}^{(i)}, \bm{z}^{(j)}) + \notag \\
   & \frac{1}{n(n-1)}\sum_{i \neq j} k(\hat{\bm{z}}^{(i)}, \hat{\bm{z}}^{(j)}) -
  \frac{2}{mn}\sum_{i, j} k(\bm{z}^{(i)}, \hat{\bm{z}}^{(j)}),
\end{align}
\endgroup
where $m$ and $n$ are the number of samples from $Z$ and $\hat{Z}$ respectively
($m$ and $n$ are batch sizes and they are equal in our experiments),
and $k\colon Z \times Z \to \mathbb{R}$ is the kernel function.
We use the information diffusion kernel \citep{NIPS2002_2216} as in W-LDA:
\begin{equation}
  k(\bm{z}, \bm{z}') = \exp (-\arccos ^2 (\sum _{i=1} ^K \sqrt{z_i z'_i})),
\end{equation}
which is sensitive to points near the simplex boundary
and thus more suitable for the sparse topic distributions.

\section{Experiments}
We evaluate our model on three datasets: 20Newsgroups consisting of 11,259 documents,
Grolier consisting of 29,762 documents, and NYTimes consisting of 99,992 documents.
We use the preprocessed 20Newsgroups of \citep{srivastava2017prodlda},
and preprocessed Grolier and NYTimes of \citep{wang2019atm}.
We compare the performance of our model with
LDA \citep{Blei2003LDA}, NVDM \citep{miao2016nvdm},
ProdLDA \citep{srivastava2017prodlda}, GraphBTM \citep{zhu-etal-2018-graphbtm},
ATM \citep{wang2019atm} and W-LDA \citep{nan-etal-2019-topic}
using topic coherence measures \citep{roder2015exploring}.
To quantify the understandability of the extracted topics,
a topic coherence measure aggregates the relatedness scores of the topic words
(top-weighted words) of each topic,
where the word relatedness scores are
estimated based on word co-occurrence statistics on a large external corpus.
For example, the NPMI coherence measure \citep{aletras-stevenson-2013-evaluating}
applies a sliding window of size 10 over the Wikipedia corpus to calculate NPMI \citep{bouma2009normalized} for word pairs.
We use three topic coherence measures in our experiments:
C\_A \citep{aletras-stevenson-2013-evaluating},
C\_P \citep{roder2015exploring},
and NPMI.
The topic coherence scores are calculated using Palmetto \citep{roder2015exploring}
\footnote{\url{https://github.com/AKSW/Palmetto}}.

\begin{table}[!hb]
  \centering
  \resizebox{\columnwidth}{!}{
    \begin{tabular}{clRRR}
      \toprule
      Dataset & Model    & \text{C\_A} & \text{C\_P} & \text{NPMI} \\
      \midrule
      \multirow{7}{*}{20Newsgroups}
              & LDA      & 0.1769      & 0.2362      & 0.0524      \\
              & NVDM     & 0.1432      & -0.2558     & -0.0984     \\
              & ProdLDA  & 0.2155      & 0.1859      & -0.0083     \\
              & GraphBTM & 0.2195      & 0.2152      & 0.0082      \\
              & ATM      & 0.1720      & 0.1914      & 0.0207      \\
              & W-LDA    & 0.2065      & 0.2501      & 0.0400      \\
              & GTM      & \bm{0.2465} & \bm{0.3451} & \bm{0.0629} \\
      \midrule
      \multirow{6}{*}{Grolier}
              & LDA      & 0.2009      & 0.1908      & 0.0498      \\
              & NVDM     & 0.1457      & -0.1877     & -0.0619     \\
              & ProdLDA  & 0.1734      & -0.0374     & -0.0193     \\
              & ATM      & 0.2189      & 0.2104      & 0.0582      \\
              & W-LDA    & 0.2354      & 0.2579      & 0.0725      \\
              & GTM      & \bm{0.2464} & \bm{0.3251} & \bm{0.0950} \\
      \midrule
      \multirow{6}{*}{NYTimes}
              & LDA      & 0.2128      & 0.3083      & 0.0773      \\
              & NVDM     & 0.1342      & -0.4131     & -0.1437     \\
              & ProdLDA  & 0.1964      & -0.0035     & -0.0282     \\
              & ATM      & 0.2375      & 0.3568      & 0.0899      \\
              & W-LDA    & 0.2253      & 0.3352      & 0.0783      \\
              & GTM      & \bm{0.2443} & \bm{0.3776} & \bm{0.0911} \\
      \bottomrule
    \end{tabular}
  }
  \caption{Average topic coherence of 5 topic number settings (20, 30, 50, 75, 100).
    Bold values indicate the best performing model under the corresponding dataset/metric setting.}
  \label{tb:avgCoherence}
\end{table}

\begin{figure*}[!ht]
  \centering
  \includegraphics[width=\textwidth]{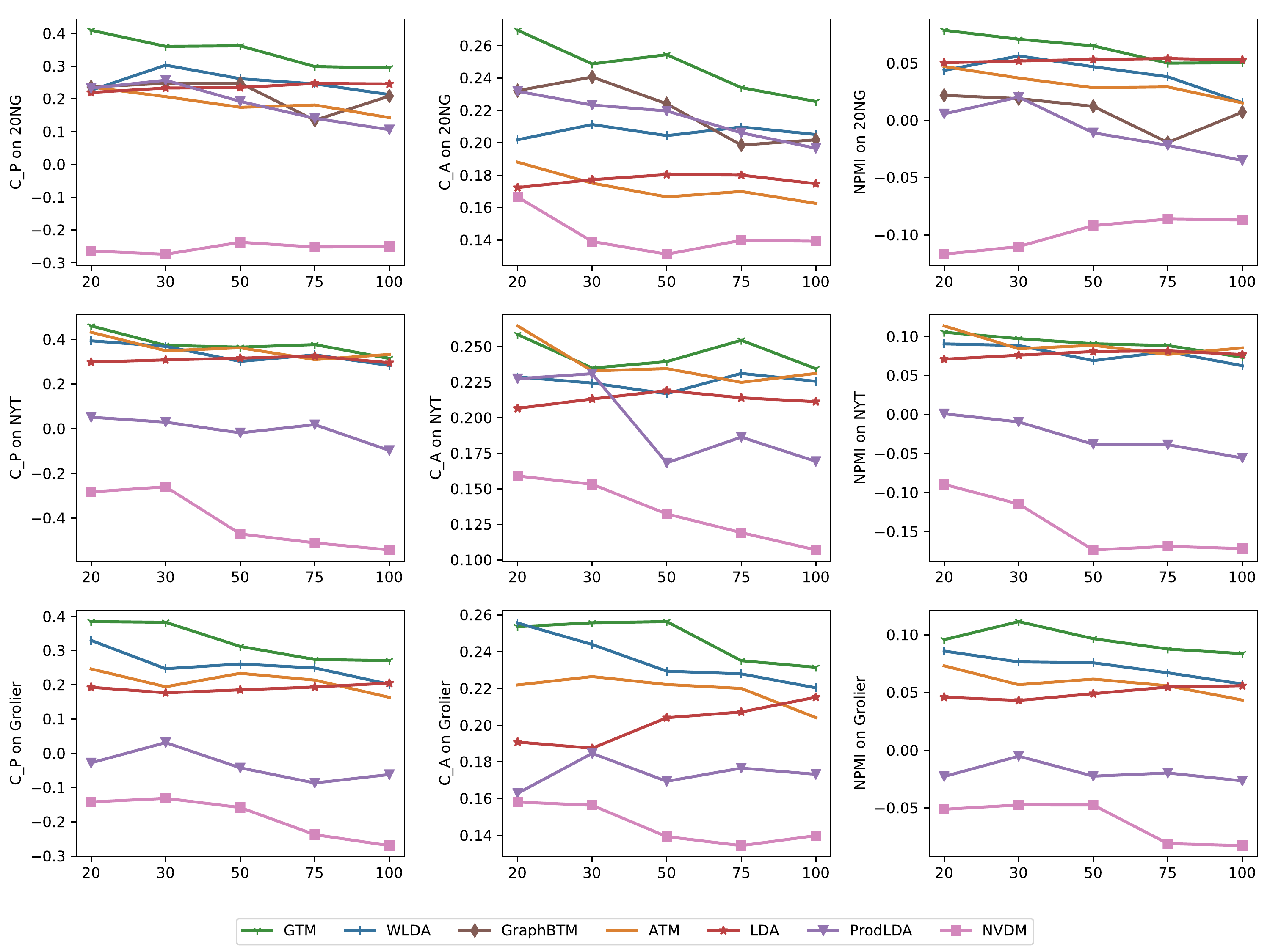}
  \caption{ Topic coherence scores (C\_P, C\_A, NPMI) w.r.t. topic numbers on 20Newsgroups (20NG), NYTimes (NYT), and Grolier.}
  \label{fig:line}
\end{figure*}

We use 2 graph convolution layers with output dimensions of $100$ and $K$ respectively in the encoder.
The hidden size of the decoder is also set to $100$.
We use the RMSProp \citep{hinton2012rmsp} optimizer with a learning rate of $0.01$ to train the model for $100$ epochs.
Since the training datasets scale up to 100K documents, i.e., 100K document nodes in the graph,
it is hard to do batch training on a single GPU given the large memory requirements.
We solve this issue by mini-batching the datasets and
feeding to the model a subgraph consisting of 1000 document nodes and all word nodes at a training step,
which results in efficient training (The training time increases almost linearly with the number of documents)
and makes it possible to apply our model to even bigger datasets.

The topic coherence results on the three datasets are shown in Table \ref{tb:avgCoherence},
where each value is the average of 5 topic number settings: 20, 30, 50, 75, 100.
From Table \ref{tb:avgCoherence}, we can observe that our proposed GTM is the best-performing model
under all dataset/metric settings.
W-LDA, ATM, LDA, and GraphBTM alternately achieve the second-best
but they are always under-performed compared to our model.
As described in section \ref{methodology},
GTM is an extension to W-LDA with the main difference that GTM models topics in a larger context
and incorporates more global information with the graph encoder.
Therefore the improvements of GTM over W-LDA indicate the effectiveness of such information for topic modeling.
We only experimented GraphBTM on 20Newsgroups
because only 20Newsgroups preserves the sequential information that is necessary for GraphBTM to build graphs.
GraphBTM performs well on the C\_A metric, which is reasonable since
C\_A is a coherence measure based on a small sliding window of size 5
and consequently prefers models concentrating on a smaller context like GraphBTM.
However, GraphBTM fails to achieve a high C\_P or NPMI score,
which uses a bigger window (70 and 10 respectively).

\begin{table*}[!ht]
  \centering
  \begin{tabular}{ll}
    \toprule
    Model & Topics                                                                                \\
    \midrule
    \multirow{3}{*}{GTM}
          & cancer medicine patient treatment medical disease md health hospital investigation    \\
          & satellite mission space launch lunar spacecraft shuttle orbit nasa flight             \\
          & car honda bmw engine ford saturn \emph{dealer} turbo rear model                       \\
          & ticket send mail price credit sale offer receive list customer                        \\
    \midrule
    \multirow{3}{*}{GraphBTM}
          & cancer hus md medical health disease patient \emph{mission} laboratory \emph{culture} \\
          & probe mission spacecraft lunar shuttle orbit nasa solar satellite space               \\
          & car bike \emph{cop} road hit gas insurance \emph{fbi guy lot}                         \\
          & \emph{car} buy \emph{mouse scsi engine} card \emph{audio pc windows faster}           \\
          & village turkish armenia azerbaijan troops militia greek lebanon armenian greece       \\
    \midrule
    \multirow{3}{*}{W-LDA}
          & msg food patient disease study science \emph{one} treatment doctor scientific         \\
          & space launch nasa satellite ground mission shuttle \emph{use} rocket orbit            \\
          & car \emph{dog} road ride speed light drive bike \emph{go} front                       \\
          & condition sale offer shipping sell \emph{excellent car speaker cd include}            \\
    \midrule
    \multirow{3}{*}{LDA}
          & \emph{use} drug cause effect medical study disease patient doctor treatment           \\
          & space launch earth nasa mission system orbit satellite \emph{design} moon             \\
          & car \emph{buy price sale new} engine \emph{offer} model \emph{dealer}                 \\
          & \emph{car} buy sell price sale \emph{new engine} offer \emph{model} dealer            \\
    \bottomrule
  \end{tabular}
  \caption{Discovered topics that are most similar to 4 ground-truth categories
    (sci.med, sci.space, rec.autos, misc.forsale)
    on 20Newsgroups with topic number 50. Italics are manually labeled off-topic words.}
  \label{tb:topics}
\end{table*}

To explore how topic coherence results vary w.r.t. different topic numbers,
we present in Figure \ref{fig:line} the topic coherence scores under different topic numbers settings.
It can be observed in Figure \ref{fig:line} that GTM enjoys the best overall performance,
achieving the highest scores in most settings.
LDA has a slightly higher NPMI score on 20Newsgroups dataset with 75 and 100 topics,
nevertheless,
GTM outperforms all baseline models with a relatively large margin on other settings of 20Newsgroups.
NVDM is apparently the worst-performing model,
while performances of models other than GTM and NVDM are not so consistent.
Notably, W-LDA, GraphBTM, and LDA obtain the second-best overall C\_P, C\_A, and NPMI scores respectively.
Another observation from Figure \ref{fig:line} is that GTM performs better on smaller topics,
probably due to the fact that
topics become more discriminative against each other when the topic number is small.

To gain an intuitive impression on the discovered topics,
we present in Table \ref{tb:topics} 4 topics corresponding to 4 out of 20 ground-truth categories of 20Newsgroups.
It can be observed that the topics discovered by GTM are more coherent and interpretable,
containing few off-topic words.
As a comparison,
GraphBTM's rec.autos topic mixes up automobiles and criminals,
W-LDA's misc.forsale topic is difficult to identify with too many off-topic words,
while LDA can not distinguish between rec.autos and misc.forsale well thus recognizes them as the same topic.
It can be observed that GTM learns more discriminative topics
by examining topic words from overlapping topics, e.g. rec.autos and misc.forsale.

\section{Conclusion}
We have introduced Graph Topic Model, a neural topic model
that incorporates corpus-level neighboring context
using graph convolutions to
enrich document representations and facilitate the topic inference.
Both quantitative and qualitative results are presented in the experiments
to demonstrate the effectiveness of the proposed approach.
In the future, we would like to extend GTM to corpora with explicit doc-doc interactions,
e.g., scientific documents with citations or social media posts with user relationships.
Replacing GCN in GTM with more advanced graph neural networks is another promising research direction.

\section*{Acknowledgments}
The authors would like to thank the anonymous
reviewers for insightful comments and helpful suggestions.
This work was funded in part by  the National Key Research and Development Program
of China (2016YFC1306704) and the National Natural Science Foundation of China (61772132).

\bibliographystyle{acl_natbib}
\bibliography{emnlp2020}

\end{document}